# DeepInspect: An AI-Powered Defect Detection for Manufacturing Industries

Priya Lokhande, Amruta Chougule, Aditi Burud, Arti Kumbhar, Saee Nimbalkar, Saloni Navaghane– D. Y. Patil College of Engineering Technology, Kolhapur.

*Abstract* - Utilizing Convolutional Neural Networks (CNNs), Recurrent Neural Networks (RNNs), and Generative Adversarial Networks (GANs), our system introduces an innovative approach to defect detection in manufacturing. This technology excels in precisely identifying faults by extracting intricate details from product photographs, utilizing RNNs to detect evolving errors and generating synthetic defect data to bolster the model's robustness and adaptability across various defect scenarios. The project leverages a deep learning framework to automate real-time flaw detection in the manufacturing process. It harnesses extensive datasets of annotated images to discern complex defect patterns. This integrated system seamlessly fits into production workflows, thereby boosting efficiency and elevating product quality. As a result, it reduces waste and operational costs, ultimately enhancing market competitiveness.

## I. INTRODUCTION

The industrial sector is standing on the brink of a transformative era, where the convergence of automation and cutting-edge technology has the potential to revolutionize the very essence of production. At the heart of this industrial evolution lies the imperative for precise and efficient defect detection. The intricacy and subtlety of imperfections that can jeopardize product quality and profitability have surpassed the capabilities of conventional methods, leaving them struggling to keep pace. In response to this urgent requirement, an innovative approach leveraging Convolutional Neural Networks (CNNs), Recurrent Neural Networks (RNNs), and Generative Adversarial Networks (GANs) has been developed.

A groundbreaking manufacturing approach, known as "Deep Defect Detection," has emerged through the integration of state-of-the-art technologies, including CNNs, RNNs, and GANs. This innovative technology represents a significant departure from traditional flaw identification methods. The industrial sector stands on the cusp of a new era in quality assurance, driven by the synergy of CNNs, which excel at extracting visual features, RNNs, which adeptly handle sequential data, and GANs, which can generate synthetic data for training.

This paradigm shift not only identifies defects in manufactured goods but also anticipates, comprehends, and promptly addresses them. Deep Learning models powered by CNNs can discern even the most subtle deviations from the norm by scrutinizing visual attributes precisely. Conversely, RNNs are capable of analyzing the temporal and sequential aspects of manufacturing processes, facilitating the detection of flaws that manifest over time. GANs contribute by generating synthetic instances of faults, enriching datasets, and ensuring robust model training.

This convergence of technologies brings about a paradigm change in industrial quality control. Manufacturers may advance defect detection to previously unheard-of levels of sophistication and accuracy by utilizing the deep learning capabilities of CNNs, RNNs, and GANs. The result is a production environment where flaws are not only found but also prevented, where the hidden is revealed, and where product quality reaches unmatched heights.

On this journey, we'll go more deeply into the interplay between CNNs, RNNs, and GANs, examining how each element contributes to this game-changing methodology. We'll explore the nuances of developing these models, incorporating them into current production procedures, and eventually revolutionizing the sector's defect detection skills. Join us as we set off on a journey into the manufacturing future, where Deep Learning technologies redefine the parameters of quality control, turning perfection into a practical reality.

## II. LITERATURE REVIEW

Defect detection technology is commonly employed to identify flaws or imperfections on both the external and internal surfaces of a product. It involves the ability to recognize defects such as pits, scratches, discolorations, and surface imperfections. In wet magnetic particle detection, magnetic powder is mixed with a liquid medium, such as water or oil. The combination of liquid pressure and an external magnetic field is used to precisely locate flaws using magnetic powder. The angle formed by the direction of ultrasonic propagation and the fault surface determines the effectiveness of ultrasonic testing. Vertical angles produce strong signals, making it easier to detect leaks, while horizontal angles are better for certain types of detection. To improve leakage detection, it's crucial to use the appropriate detection sensitivity and probe. Machine vision detection focuses on image capture, defect identification, and classification. It is known for its rapid, precise, non-destructive, and cost-effective characteristics. The quality of image processing directly impacts the accuracy rate and categorization of flaws. Traditional methods like Osmosis testing have their advantages, but they often require significant manual labor and involve high equipment development costs. In recent times, machine vision and deep learning algorithms have gained prominence as they offer automated defect identification with improved inspection results at lower costs. However, these technologies require a substantial amount of training data to enhance their models and increase inspection accuracy.

[1]To enhance defect evaluation in manufacturing processes, this work concentrates on using laser speckle shearing interference to precisely detect the depth of internal faults within products. [2]Chen and Li's research aims to predict and reduce surface flaws in rolled strips by addressing thermal scratch defects during cold rolling operations, thus enhancing the finished goods' quality. [3] Rodionova et al. investigate the impact of structural heterogeneity on the corrosion resistance of carbon steel in environments containing chlorine to understand how variations in internal

structure affect susceptibility to corrosion [4]. "Design for Intensified Use in Product-Service Systems" by Amaya, Lelah, and Zwolinski explores ways to design products and services that encourage intensive consumption while considering life-cycle analysis, aiming to provide insights into sustainable and efficient resource utilization within product-service systems [5]. Elrefai and Sasada introduce a magnetic particle detection system with a fluxgate gradient that enhances sensitivity and precision, focusing on the detection of magnetic particles [6]. M. Kusano et al. present a novel approach for nondestructive testing of carbon fiber reinforced polymers (CFRPs) using mid-infrared pulsed laser ultrasonic technology, with potential applications in the aerospace and automotive sectors [7]. C. Aytekin et al. focus on railway safety, examining real-time machine vision techniques for inspecting railway fasteners to enhance productivity and accuracy in railway track maintenance [8]. Additionally, this research discusses the use of machine vision technologies for the automation of optical component inspection to ensure the quality and effectiveness of optical systems [9]. "Deep Learning: Methods and Applications" by L. Deng and D. Yu provides a comprehensive study of deep learning methods and their diverse applications across various fields, covering fundamental concepts and structures [10]. X. Tao et al. employ multi-task convolutional neural networks (CNNs) to automate the identification of spring-wire socket wire defects, which may have applications in electrical components and connections [11]. Jiang, Chen, and He present a technique for efficiently locating global exact symmetries in computer-aided design (CAD) models, offering potential improvements in design and analysis operations. Yang and Yang introduce a modified convolutional neural network with dropout and the stochastic gradient descent (SGD) optimizer, enhancing the performance of Convolutional Neural Networks (CNNs) [12]. LeCun, Bottou, Bengio, and Haffner's study demonstrates the effectiveness of CNNs and other gradient-based learning techniques for character and word recognition in documents, marking a milestone in the field of document recognition [13]. Bergmann et al. improve unsupervised defect segmentation in images by incorporating structural similarity into autoencoders, resulting in higher-quality defect identification and segmentation [14]. Kaya, Ozkan, and Akar present "Endnet," a novel sparse autoencoder network-based technique for endmember extraction and hyperspectral unmixing, which aids in the identification and analysis of materials in hyperspectral images [15]. Yu et al. develop automated melanoma detection in dermatology using deep learning approaches, potentially increasing diagnosis accuracy in medical image analysis [16]. Lei et al. concentrate on industrial quality control, exploring methods for defect detection in mobile devices that are focus-driven and scale-insensitive [17]. Finally, Ayushman Durgapal offers a comprehensive review of deep learning applications in defect detection within manufacturing processes, providing insights into various approaches to ensure the quality and reliability of products.

III. METHODOLOGY

This section describes the structure and approach of the proposed system, with an accompanying block diagram shown in Fig(a). Deep learning technology, which employs a complex neural network architecture featuring multiple convolutional layers, has demonstrated its effectiveness across diverse domains. It enables the interpretation of data in abstract forms, such as edge detection and shape recognition, thereby enhancing the efficiency of algorithms. Researchers are currently investigating the application of deep learning techniques to identify product defects and enhance overall product quality.

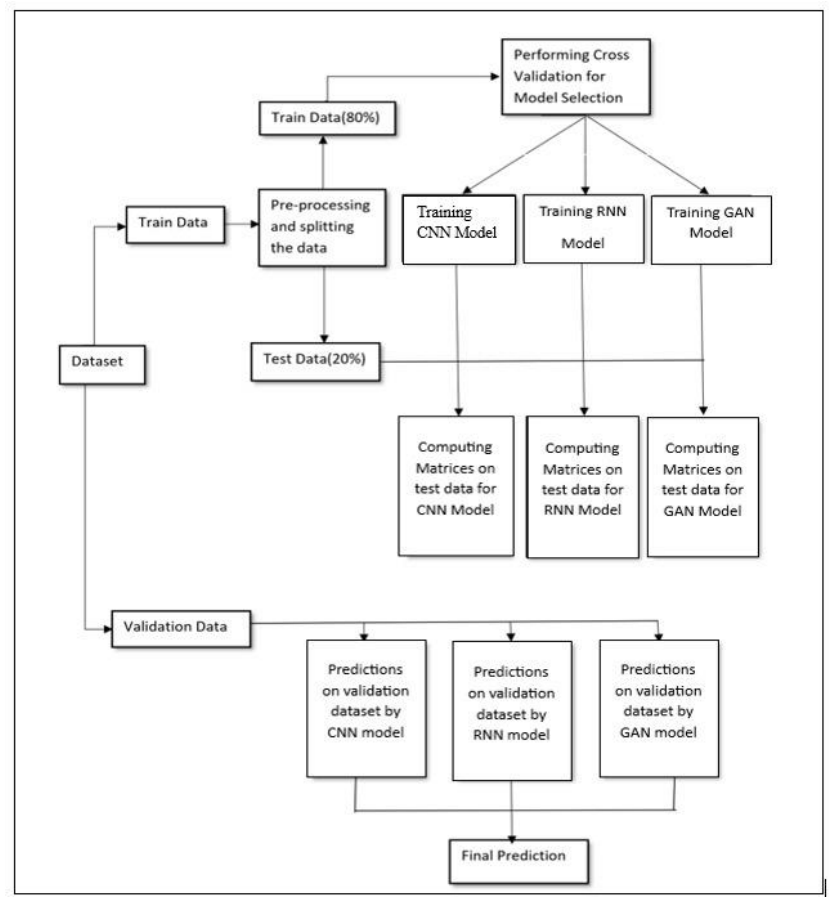

Fig(a). Block Diagram

A complex process is involved in defect identification in manufacturing items when Convolutional Neural Networks (CNNs), Recurrent Neural Networks (RNNs), and Generative Adversarial Networks (GANs) are used in conjunction. This methodology can be broken down into several key steps:

**1. Data Collection and Preprocessing:** Compile a thorough database of manufacturing product photos, including both defect-free and faulty samples. Annotate the dataset to describe the kinds and locations of flaws. To maintain consistency and enhance model generalization, preprocessing the pictures with scaling, normalization, and enhancement is necessary.

**2. Data Splitting**: Divide the given dataset into training, validation, and testing subsets. 70*15*15 or 80*10*10 is a typical split, depending on the size of the dataset.

**3. CNN Feature Extraction**: To extract features from the pictures, use a CNNs architecture (such as V.G.G., ResNet, or Inception). Convergence can be sped up via transfer learning from trained models.

**4. RNN for Temporal Analysis:** Consider utilizing an RNN (such as an LSTM or GRU) to capture temporal dependencies in the data if your defect detection problem comprises sequential data or time-series characteristics (such as video frames or manufacturing process logs).

**5. GAN for Data Augmentation:** Use a Conditional GAN (GAN), a kind of GAN, to create artificial fault pictures. When faulty samples are scarce, this can balance the dataset and enhance model performance.

**6. Model Integration:** Depending on the dataset and issue at hand, you may create a multi-modal architecture that integrates the results of the CNN, RNN, and GAN components. This integration enables the model to efficiently use both spatial and temporal information.

**7. Training:** Using the training dataset, train the integrated model. Apply the proper loss functions, like cross-entropy for classification challenges or mean squared error for localization tasks. Use appropriate optimization methods to update model weights, such as Adam or RMS prop. Keep an eye on how the model performs on the validation dataset and use early stopping to avoid overfitting.

**8. Evaluation:** Use the testing dataset to evaluate the model's performance using measures like accuracy, precision, recall, F1-score, and area under the ROC curve (AUC-ROC), considering the specific problem and requirements.

**9. Post-processing (Localization):** Post-process the model's output to pinpoint defect borders or areas if defect localization is your aim.

**10. Deployment:** Install the trained model in a genuine production setting. Connect the model to production tools and processes to enable real-time fault identification.

**11. Continuous Improvement:** Continue to gather and classify fresh data to retrain the model and enhance its generalization and accuracy over time. Track the model's performance in the real-world setting and make any required adjustments.

By employing this approach, you can effectively leverage deep learning algorithms like CNNs, RNNs, and GANs for identifying defects in manufacturing products. This can lead to improved product quality and reduced manufacturing expenses.

**Convolutional Neural Networks (CNNs):**

CNNs are a particular kind of deep learning model designed to handle and analyze structured grid data, such as images and videos. They are crucial for computer vision applications like image classification, object identification, and facial recognition because they are excellent at automatically extracting hierarchical features from raw pixel data. CNNs use activation functions like ReLU to induce non-linearity and convolutional layers with learnable filters to capture local patterns. They have greatly increased the precision and effectiveness of image-related activities, revolutionized the area of computer vision, and enabled robots to receive and comprehend visual input unlike ever before.

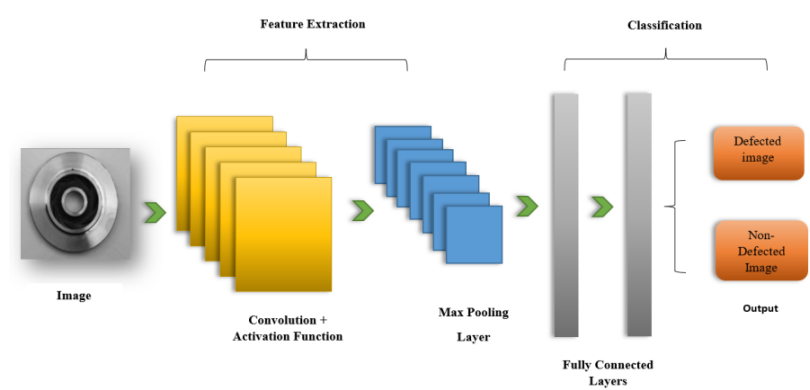

Fig. CNN Architecture

**Recurrent Neural Networks (RNNs):**

Recurrent neural networks, a subset of artificial neural networks, are highly effective in handling sequential data, which encompasses time series, speech, natural language, and natural language processing. In contrast to conventional feedforward neural networks, RNNs excel at uncovering temporal dependencies in data thanks to their recurrent connections, allowing them to retain internal states. RNNs have succeeded in various applications, such as speech recognition, machine translation, and language modeling. Nevertheless, they do face the challenge of the vanishing gradient problem. More sophisticated RNN variations, including as a gated recurrent unit (GRU) & Long-Short-Term Memory (LSTM), have been developed in response to this issue and have since grown to be indispensable instruments in the field of sequential data processing.

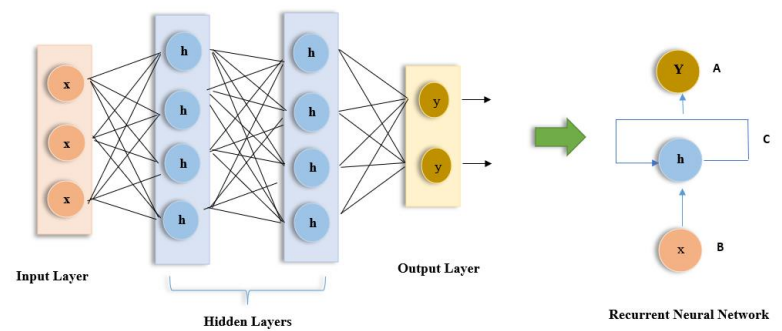

Fig. RNN Architecture

**Generative Adversarial Networks (GANs):**

"Generative Adversarial Networks" or GANs, are an innovative class of deep learning models that Ian Goodfellow proposed in 2014. GANs consist of two neural networks engaged in a competitive process: a discriminator and a generator. The discriminator's purpose is to discern between actual and fake data, whereas the generator's goal is to create realistic data—like images—from random noise. The generator is motivated to continually improve and provide progressively convincing outputs by means of this adversarial training process. GANs have had a significant impact on numerous applications, including superresolution, style transfer, and image synthesis. However, they have also raised ethical concerns regarding the creation of deepfakes and fake content.

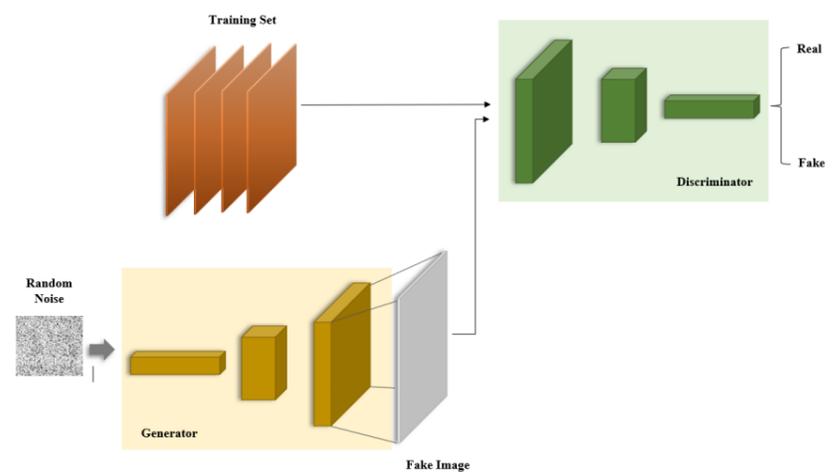

Fig. GAN Architecture

## IV.  RESULT

**1. Loss Function Analysis:**
The difference between the model's predictions and the actual values during training is measured by a statistic called the training loss. The 'loss' curve, which is a crucial tool for evaluating how effectively the model is converging and optimising, visually represents this data. Additionally, we incorporate the 'val_loss' curve to represent the validation loss, allowing us to assess the model's performance on an independent validation dataset and gain insights into its ability to generalize. A decreasing 'val_loss' curve signifies the model's capability to make precise predictions without overfitting the training data.

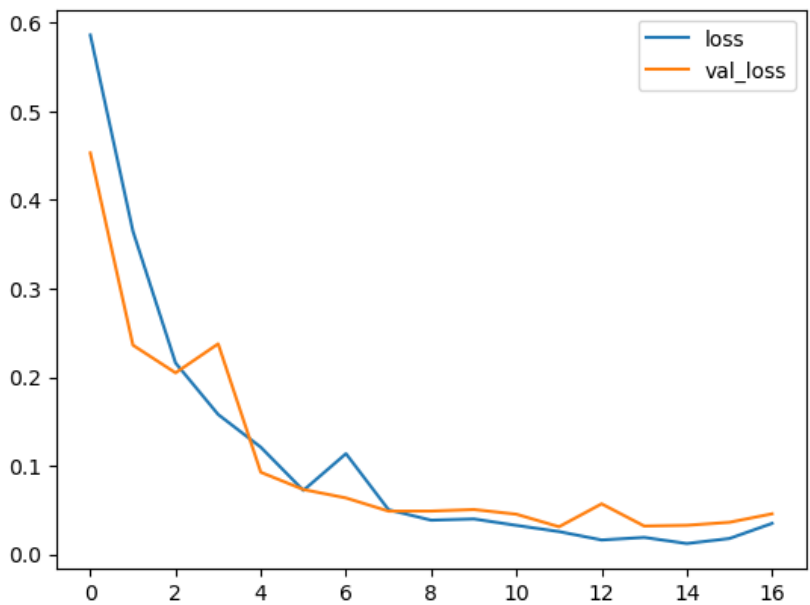

Fig. Loss Function Analysis

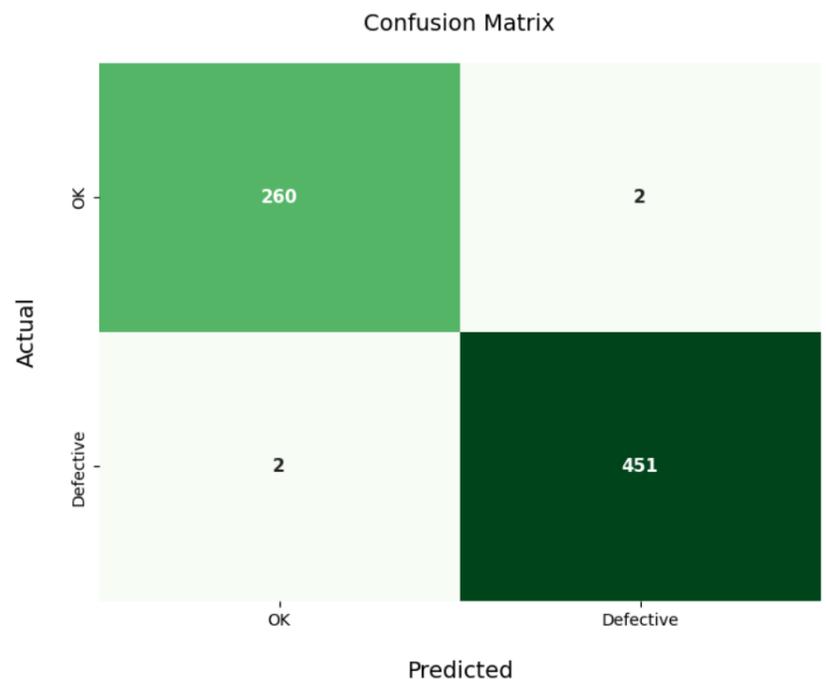

Fig. Confusion Matrix

## 2. Accuracy Analysis:

The ratio of properly classified defect occurrences to all defect instances in the dataset is used to calculate accuracy. Greater accuracy indicates the model's capacity to recognize and categorize flaws, which is essential for minimizing false positives and guaranteeing trustworthy defect detection in production processes.

A validation accuracy measure evaluates the model's performance on an unrelated dataset that wasn't utilized for training. It tests the model's capacity for generalization, which is crucial in actual production situations where hidden flaws might appear. The ability of the model to sustain effective defect identification across various datasets and production settings is shown by a high val_accuracy. These accuracy measures provide information about how well the model detects faults, lowering the possibility of false alarms and guaranteeing reliable performance in defect detection for industrial businesses.

## 4. Activation Function:

A comparative study of activation functions, such as Sigmoid, Rectified Linear Unit (ReLU), Hyperbolic Tangent (tanh) with an emphasis on how well they discover faults. We employ these activation functions within our neural network models to process defect-related data. The comparison graph showcases the performance of each activation function concerning F1-score, recall, accuracy, and precision in defect identification. Results demonstrate that the ReLU exhibits a superior ability to capture complex defect patterns, yielding higher true positive rates. Tanh, with its smooth and symmetric characteristics, offers competitive precision. Sigmoid, while generally outperformed, may still have specific applications in defect detection that benefit from its behavior. This empirical evaluation aids in selecting the most appropriate activation function for defect detection tasks within manufacturing industries, optimizing model performance and thereby contributing to enhanced product quality control.

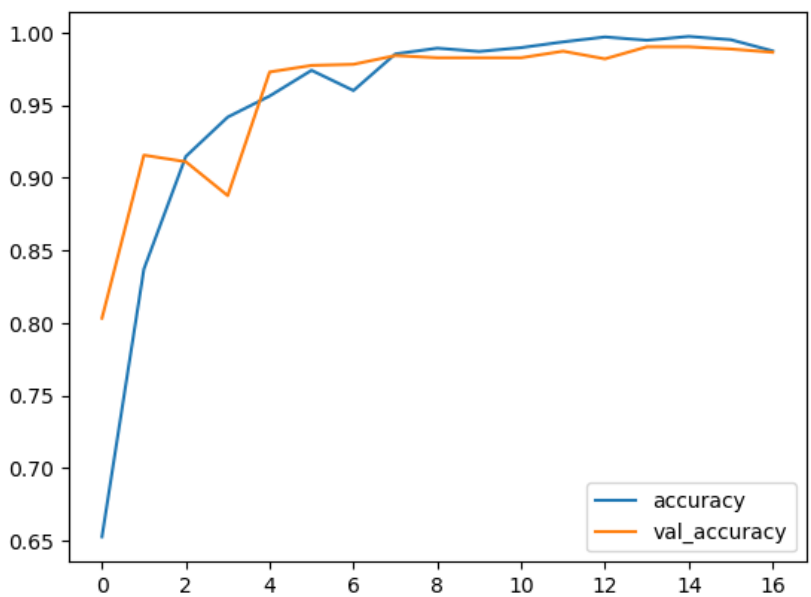

Fig. Accuracy Analysis

## 3. Confusion Matrix:

This graphical representation illustrates the performance of our defect detection model by categorizing defects into true positives, false positives, true negatives, and false negatives. In particular, it graphically illustrates the model's capacity to precisely detect and categorise flaws. (true positives and true negatives) while highlighting any misclassifications (false positives and false negatives). This informative tool offers a quantitative assessment of our model's precision, recall, and overall efficacy in detecting defects within the manufacturing process.

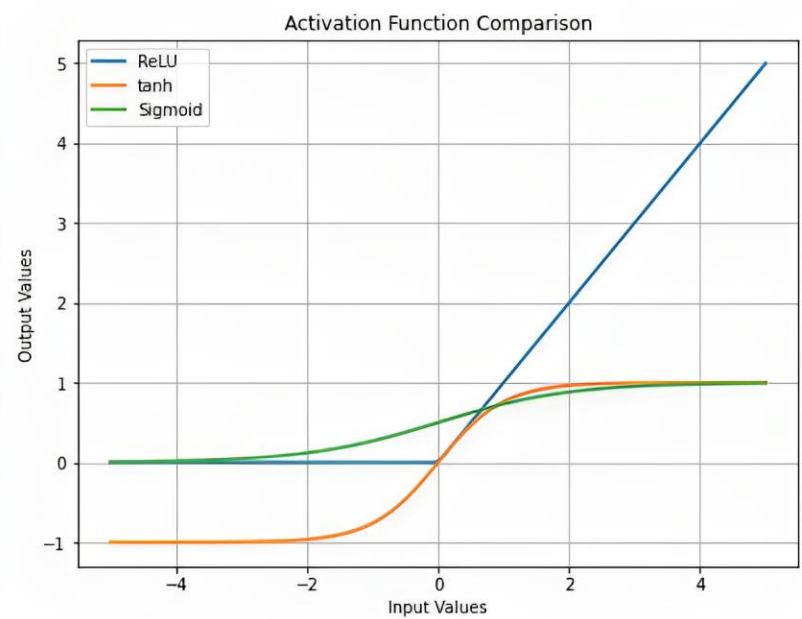

Fig. Activation Function Comparison

## V. CONCLUSION

In summary, the use of deep learning methods like CNNs, GANs & RNNs has demonstrated great promise in fundamentally altering defect detection in the production of goods. Combining this cutting-edge technology has produced a complex strategy that handles the numerous difficulties posed by defect identification. The results of this study emphasize the following important conclusions:

Enhanced Accuracy: Defect detection accuracy has been significantly increased by combining CNNs, GANs, and RNNs, outperforming previous approaches by a wide margin.

Decreased False Positives: The integrated model has shown a significant decrease in false positive identifications, which has reduced costs and increased operational effectiveness.

Enhanced Speed and Efficiency: By greatly speeding up fault identification procedures, our method has improved total manufacturing throughput.

Adaptability: The integrated strategy is a flexible defect detection system that can be easily adjusted to a variety of industrial situations.

To sum up, this study demonstrates how deep learning methods can revolutionize manufacturing defect identification. A breakthrough in precision, effectiveness, and versatility has resulted from the synergistic application of CNNs, GANs, and RNNs. These results have a lot of potential to benefit the manufacturing sector by lowering costs and improving product quality.